\documentclass[10pt,twocolumn,letterpaper]{article}
\usepackage{cvpr}
\usepackage[USenglish]{babel}
\usepackage{tikz}
\usetikzlibrary{positioning,calc,matrix,arrows}
\usepackage{times}
\usepackage{graphicx}
\usepackage{amsmath}
\usepackage{amssymb}
\usepackage{booktabs}
\usepackage{xcolor}
\usepackage{array}
\usepackage{bm}
\usepackage{mathtools}
\usepackage[breaklinks=true,bookmarks=false]{hyperref}

\tikzstyle{block} = [draw, rectangle, minimum height=3em, minimum width=3em]
\tikzstyle{data} = []
\tikzstyle{datac} = [draw, circle, minimum height=1em, minimum width=1em,inner sep=3pt]
\tikzstyle{par} = [draw, circle, minimum height=1em, minimum width=1em,fill=black!20,inner sep=3pt]
\tikzstyle{pinstyle} = [pin edge={to-,thin,black}]
\tikzstyle{to} = [->,>=stealth',shorten >=1pt,semithick]

\newcolumntype{L}[1]{>{\raggedright\let\newline\\\arraybackslash\hspace{0pt}}m{#1}}
\newcolumntype{R}[1]{>{\raggedleft\let\newline\\\arraybackslash\hspace{0pt}}m{#1}}
\newcolumntype{C}[1]{>{\centering\let\newline\\\arraybackslash\hspace{0pt}}m{#1}}

\newcolumntype{x}{>\small c}
\newcommand{\bx}{\mathbf{x}}

\newcommand{\bw}{\mathbf{w}}
\newcommand{\bs}[1]{\boldsymbol{#1}}
\newcommand{\argmin}{\operatornamewithlimits{argmin}}

\renewcommand{\paragraph}[1]{\par\medskip\noindent\textbf{#1}}

\cvprfinalcopy
\def\cvprPaperID{XXX} 

\ifcvprfinal\fi

\title{%
Universal representations:\\
The missing link between faces, text, planktons, and cat breeds}

\author{%
Hakan Bilen\\
University of Oxford\\
{\tt\small hbilen@robots.ox.ac.uk}
\and
Andrea Vedaldi\\
University of Oxford\\
{\tt\small vedaldi@robots.ox.ac.uk}
}

\begin{document}
\maketitle
\begin{abstract}
With the advent of large labelled datasets and high-capacity models, the performance of machine vision systems has been improving rapidly. However, the technology has still major limitations, starting from the fact that different vision problems are still solved by different models, trained from scratch or fine-tuned on the target data. The human visual system, in stark contrast, learns a \emph{universal representation} for vision in the early life of an individual. This representation works well for an enormous variety of vision problems, with little or no change, with the major advantage of requiring little training data to solve any of them.

In this paper we investigate whether neural networks may work as universal representations by studying their capacity in relation to the ``size'' of a large combination of vision problems. We do so by showing that a single neural network can learn simultaneously several very different visual domains (from sketches to planktons and MNIST digits) as well as, or better than, a number of specialized networks. However, we also show that this requires to carefully normalize the information in the network, by using domain-specific scaling factors or, more generically, by using an instance normalization layer.
\end{abstract}

\section{Introduction}\label{s:intro}

\begin{figure}
\begin{center}
\includegraphics[width=\columnwidth]{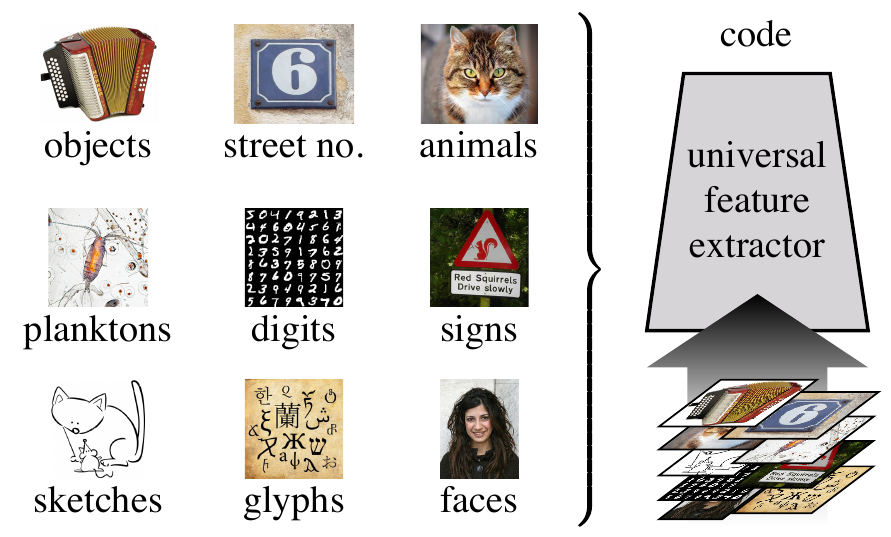}
\end{center}
\caption{Humans posses an internal visual representation that, out of the box, works very well any number of visual domains, from objects and faces to planktons and characters. In this paper we investigate such {\bf universal representations} by constructing neural networks that work simultaneously on many domains, learning to share common visual structure where no obvious commonality exists. Our goal is to contrast the capacity of such model against the total size of the combined vision problems.}\label{f:splash}
\end{figure}

While the performance of machine vision systems is nowadays believed to be comparable or even superior to the one of human vision in certain tasks~\cite{he15delving}, the very narrow scope of these systems remains a major limitation. In fact, while vision in a human works well for an enormous variety of different problems, different neural networks are required in order to recognize faces~\cite{hong16faces,parkhi15deep,schroff15facenet:,taigman14deepface:}, classify~\cite{krizhevsky12imagenet, he16deep}, detect~\cite{ren15faster,liu16ssd:} or segment~\cite{dai16instance-aware} high-level object categories, read text~\cite{gupta16synthetic,jaderberg16reading}, recognize bird, flower, dog, or cat species~\cite{lin15bilinear}, interpret a radiography, an echography, or a MRI image of different parts of the human anatomy~\cite{jamaludin16spinenet:}, and so on. 

Differently from machines, humans develop a powerful internal representation of images in the early years of their development~\cite{atkinson02the-developing}. While this representation is subject to slight refinements even later in life, it changes little. This is possible because the representation has a \emph{universal valence} and works equally well for any number of problems, from reading text to recognizing people and contemplating art.

The existence of non-trivial general-purpose representations means that an significant part of vision can essentially be learned once for all. However, the nature and scope of such universal representations remains unclear. In this paper, we shed some light on this question by investigating to which extent deep neural networks can be shared between extremely diverse visual domains (Fig.~\ref{f:splash}).

We start our investigation by asking whether it is possible to learn neural networks simultaneously from a large number of different problems (Fig.~\ref{f:splash}). Several authors~\cite{oquab14learning,razavian14cnn-features,yosinski14how-transferable} have shown that neural networks can \emph{transfer} knowledge between tasks through a process of adaptation called \emph{fine tuning}. While this is encouraging, here we look at the much more challenging problem of learning a single network that works well for \emph{all the problems simultaneously}.

Several authors have considered multi-task scenarios before us, where the task is to extract multiple labels from the same visual domain (\eg image classification, object and part detection and segmentation, and boundary extraction, all in PASCAL VOC~\cite{dai16instance-aware,bilen16integrated,kokkinos2016ubernet}). Such tasks are expected to partially overlap since they look at the same object types. Our goal, instead, is to check whether extreme visual diversity still allows a sharing of information. In order to do so, we fix the labelling task to image classification, and look at combining numerous and diverse domains (\eg text, faces, animals, objects, sketches, planktons, etc.).

While the setup is simple, it allows to investigate an important question: what is the capacity of models in relation to the ``size'' of the combination of multiple vision problems. If problems are completely independent, the total size should grow proportionally, which should be matched by an equally unbounded increase in model capacity. On the other hand, if problems overlap, then the complexity growth gradually slows down, allowing model complexity to catch up, so that, in the limit, universal representations become possible.

Our first contribution is to show, through careful experimentation (section~\ref{s:experiments}), that the capacity of neural networks is large even when contrasted to the complexity generated by combining numerous and very diverse visual domains. For example, it is possible to share \emph{all layers of a CNN, including classification ones}, between datasets as diverse as CIFAR-10, MNIST and SVHN, without loss in performance (section~\ref{s:small}). In general, extensive sharing of parameters works very well for combination of up to ten diverse domains.

Our second contribution is to show that, while sharing is possible, it notelessly requires to normalize information carefully, in order to compensate for the different dataset statistics (section~\ref{s:method}). We test various schemes, including domain-oriented batch and instance normalization, and find (section~\ref{s:experiments}) that the best method uses domain-specific scaling parameters learned to compensate for the statistical differences between datasets. However, we also show that instance normalization can be used to construct a representation that works well for all domains while using a single set of parameters, without any domain-specific tuning at all.

\section{Related Work}\label{s:related}

\paragraph{Transfer learning and domain adaptation.} Our work is related to methods that transfer knowledge between different tasks and between same tasks from different domains. Long \etal~\cite{long2015learning} propose the Deep Adaptation Network, a multi-task and multi-branch network that matches hidden representations of task-specific layers in a reproducing kernel Hilbert space to reduce domain discrepancy. Misra \etal~\cite{misra2016cross} propose Cross-Stitch units that combine the activations from multiple networks and can be trained end-to-end. Ganin and Lempitsky~\cite{ganin2015unsupervised} and Tzeng \etal~\cite{tzeng2015simultaneous} propose deep neural networks that are simultaneously optimized to obtain domain invariant hidden representations by maximising the confusion of domain classifiers. Yosinski \etal~\cite{yosinski14how-transferable} study transferability of features in deep neural networks between different tasks from a single domain. The authors investigate which layers of a pre-trained deep networks can be adapted to new tasks in a sequential manner. The previous work explore various methods to transfer between different networks, here we look at learning universal representations from very diverse domains with a single neural network.

Our work is also related to methods \cite{hinton2015distilling,romero2014fitnets,chen2015net2net} that transfer the information between networks. Hinton \etal~\cite{hinton2015distilling} propose a knowledge distillation method that transfers the information from an ensemble of models (teacher) to a single one (student) by enforcing it to generate similar predictions to the existing ones. Romero \etal~\cite{romero2014fitnets} extend this strategy by encouraging similarity between not only the predictions but also between intermediate hidden representations of different networks. Chen~\etal~\cite{chen2015net2net} address the slow process of sequential training both teacher and student networks from scratch. The authors accelerate the learning process by simultaneously training teacher and student networks. This line of work focuses on learning compact and accurate networks on the same task by transferring knowledge between different networks, while our work aims to learn a single network that can perform well in multiple domains.
 
\paragraph{Multi-task learning.} Multi-task learning~\cite{caruana97multitask} has been extensively studied over two decades by the machine learning community. It is based on the key idea that the tasks share a common representation which is jointly learnt along with the task specific parameters. Multi-task learning is applied to various computer vision problems and reported to achieve performance gains in object tracking~\cite{Zhang13robust}, facial-landmark detection \cite{zhang14facial}, surface normals and edge labels~\cite{wang2015designing}, object detection and segmentation~\cite{Dai16instance} object and part detection~\cite{bilen16integrated}. In contrast to our work, multi-task learning typically focuses on different tasks in the same datasets.

\paragraph{Life-long learning.} Never Ending Learning~\cite{mitchell2010never} and Life-long Learning~\cite{thrun1998lifelong,silver2013lifelong} aim at learning many tasks sequentially while retaining the previously learnt knowledge. Terekhov~\etal~\cite{terekhov2015knowledge} propose Deep Block-Modular Neural Networks that allow a previously trained network learn a new task by adding new nodes while freezing the original network parameters. Li and Hoiem~\cite{li2016learning} recently proposed the Learning without Forgetting method that can learn a new task while retaining the responses of the original network on the new task. The main focus in this line of research is to preserve information about old tasks as new tasks are learned, while our work is aimed at exploring the capacity of models when multiple tasks are learned jointly.

\section{Method}\label{s:method}

\begin{figure*}[t]
\centering\includegraphics[width=\textwidth]{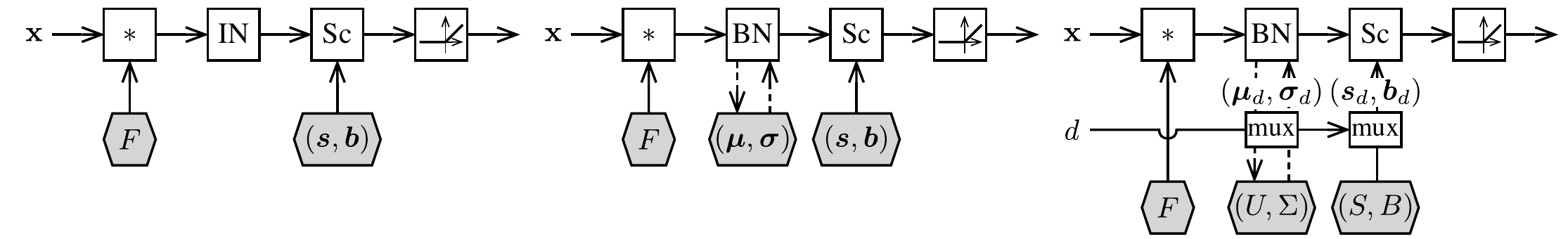}
\caption{From left to right, three example modules: instance normalization, batch normalization, and batch normalization with domain-specific scaling building modules. The shaded blocks indicate learnable parameters. Other variants are tested, not shown for compactness.}\label{f:arch}
\end{figure*}

We call a \emph{representation} a vectorial function $\phi: \bx \mapsto \phi(\bx)\in\mathbb{R}^C$ mapping an image $\bx \in \mathbb{R}^{H\times W \times 3}$ to a $C$-dimensional code vector $\phi(x)$ (often this vector is also a 3D tensor). As representations we consider here deep convolutional neural networks (DCNNs). A DCNN can be decomposed as a sequence $\phi(\bx)=\phi_N \circ \dots \circ \phi_2 \circ\phi_1(\bx)$ of linear and non-linear functions $\phi_n$, called \emph{layers}, or, in more sophisticated cases~\cite{szegedy16rethinking,he16deep}, as a feed-forward computational graph where such functions are used as nodes.

A concept that will be important later is the one of \emph{data batches}. Neural networks are almost invariably learned by considering batches of example images together. Here we follow the standard practice of representing a batch of $T$ images by adding a fourth index $t=1,\dots,T$ to the data tensors $\bx \in \mathbb{R}^{H\times W \times 3 \times T}$. As the information propagates through the network, all intermediate tensors are also batches of $T$ data points.

\subsection{Learning from multiple domains}\label{s:mdomain}

Here we consider the problem of learning neural networks from multiple domains $\mathcal{D}_1,\dots,\mathcal{D}_D$. For simplicity, we limit ourselves to \emph{image classification} problems. Hence, a domain $\mathcal{D}_d = (\mathcal{X}_d,\mathcal{Y}_d,p_d,\mathcal{L}_d)$ consists of an input space $\mathcal{X}_d=\mathbb{R}^{H_d\times W_d\times C_d}$, a discrete label (output) space $\mathcal{Y}_d =\{1,2,\dots,K_d\}$, an (unknown) joint probability distribution $p_d(\bx,y)$ over inputs and labels, and a loss function $\mathcal{L}_d:\mathcal{Y}\times\mathcal{Y}\rightarrow \mathbb{R}$ measuring the quality $\mathcal{L}_d(y,\hat y)$ of a label prediction $\hat y$ against its ground truth value $y$. As usual, the quality of a predictor $\hat y = \phi_d(\bx)$ is measured in terms of its expected risk $E[\mathcal{L}_d(y,\phi_d(\bx))]$. For each domain, furthermore, we also have a training set $\mathcal{T}_d=\{(\bx^{(1,d)},y^{(1,d)}),\dots,(\bx^{(N_d,d)},y^{(N_d,d)})\}$ of $N_d$ training pairs, which results in the empirical risk
\[
 V_d(\phi_d) = \frac{1}{N_d}\sum_{t=1}^{N_d} \mathcal{L}_d(y^{(t,d)},\phi_d(\bx^{(t,d)}))
\]
We also assume that a similar but disjoint validation set $\mathcal{V}_d$ is available for each domain.

Our goal is to learn $D$ predictors $\phi_1,\dots,\phi_D$, one for each task, in order to minimize their overall risk. While balancing different tasks is an interesting problem in its own right, here we simply choose to minimize the average risk across domains:
\begin{multline}\label{e:objective}
(\phi_1^*,\dots,\phi_D^*)
=\\
\argmin_{\phi_1,\dots,\phi_D}
\lambda \mathcal{R}(\phi_1,\dots,\phi_D)
+ \frac{1}{D} \sum_{d=1}^{D} V_d(\phi_d).
\end{multline}
The term $\mathcal{R}(\phi_1,\dots,\phi_D)$ encodes both regularization terms as well as hard constraints, defining the structure of the learning problem.

\paragraph{No sharing.} As a baseline, separate neural networks are learned for each domain. This is obtained when the regularizer in Eq.~\ref{e:objective} decomposes additively $\mathcal{R}(\phi_1,\dots,\phi_D) = \sum_{d=1}^D \mathcal{R}(\phi_d)$. In this case, there is \emph{no sharing} between domains.

\paragraph{Feature sharing.} The baseline is set against the case in which part of the neural networks $\phi_d$ are shared. In the simplest instance, this means that one can write
\[
   \phi_1 = \phi_1'\circ\phi_0,
   \quad\dots,
   \quad\phi_D = \phi_D'\circ\phi_0,
\]
where $\phi_0$ is a common subset of the networks. For example, following the common intuition that early layers of a neural networks have are less specialized and hence less domain-specific~\cite{cimpoi14deep}, $\phi_0$ may contain all the early layers up to some depth, after which the different networks branch off.\footnote{Such a constraint can be incorporated in~Eq.~\ref{e:objective} by requiring that $R(\phi_1,\dots,\phi_d) < \infty~\Leftrightarrow~\exists \phi_0:\ \forall d=1,\dots,D\ \exists \phi_d':\ \phi_d = \phi_d'\circ \phi_0$.} We call this \emph{ordinary feature sharing}.

\paragraph{Adapted feature sharing.} In this paper, we propose and study alternatives to ordinary feature sharing. More abstractly, we are interested in minimizing the difference between the individual representations $\phi_1,\dots,\phi_D$ and a universal representation $\phi_0$ used as a common blueprint. For example, when domains differ substantially in their statistics (\eg text vs natural images), such differences may have a significant impact in the response of neurons, but it may be possible to compensate for such differences by slightly adjusting the representation parameters. Another intuition is that not all features in the universal representation $\phi_0$ may be useful in all domains, so that some could be deactivated depending on the problem. In order to explore these ideas, we will consider the case in which representations decompose as $\phi_d(\bx) = \phi_d' \circ \phi_0(\bx|\bw_d)$, where $\bw_d$ is a small number of domain-dependent parameters and $\phi_0(\cdot;\bw_d)$ is the universal representation blueprint. We call this \emph{adapted feature sharing}.

For adapted feature sharing, we consider in particular an extremely simple form of parametrization for $\phi_0(\cdot;\bw_d)$ (Fig.~\ref{f:arch}). In order to be able to adjust for different mean responses of neurons for each domain, as well as to potentially select subset of features, we consider adding a domain-dependent \emph{scaling factor}  $\bm s_d$ and a \emph{bias} $\bm b_d$ after each convolutional or linear layer in a CNN. This is implemented by a \emph{scaling layer} $\phi_\text{scale}$:
\[
\forall vuc:\ y_{vuc} = [\phi_\text{scale}(\bx;\bm s,\bm b)]_{uvc}
=
s_c\,x_{vuc} + b_c.
\]
All together, the scale and bias parameters form collections $S=(\bm s_1,\dots,\bm s_D)$ and $B=(\bm b_1,\dots,\bm b_D)$. Since all domains are trained jointly, we introduce also a \emph{muxer} (Fig.~\ref{f:arch}), namely a layer that extracts the corresponding parameter set given the index $d$ of the current domain:
\[
  \phi_\text{mux}(d;S,B) = (\bm s_d,\bm b_d).
\]

For networks that include batch or instance normalization layers (Section~\ref{s:norm}), a scaling layer already follows each occurrence of such blocks. In this case, we simply adapt the corresponding parameters rather than introducing new scaling layers.

\subsection{Batch and instance normalization}\label{s:norm}

\emph{Batch normalization} (BN)~\cite{ioffe15batch} is a simple yet powerful technique that can substantially improve the learnability of deep neural networks. The batch normalization layer is defined as
\begin{equation}\label{e:bnorm}
y_{vuct}
=
[\phi_{\text{BN}}(\bs\bx)]_{vuct}
=
\frac{x_{vuct} - \mu_c(\bx)}{\sqrt{\sigma_c^2(\bx)+\epsilon}}.
\end{equation}
where the batch means $\bs\mu$ and variances $\bs\sigma^2$ are given by
\begin{align*}
 \mu_{c}(\bx) &= \frac{1}{HWT}\sum_{vut} x_{vuct},\\
 \sigma^2_{c}(\bx) &= \frac{1}{HWT}\sum_{vut} (x_{vuct} - \mu_c)^2.
\end{align*}

Recently,~\cite{ulyanov16instance,ba16layer} noted that it is sometimes advantageous to simplify batch normalization further, and consider instead instance (or layer) normalization. The \emph{instance normalization} (IN) layer, in particular, has almost exactly the same expression as~Eq.~\ref{e:bnorm}, but mean and covariance are instance rather than batch averages:
\begin{equation}\label{e:inorm}
y_{vuct}
=
[\phi_{\text{IN}}(\bs\bx)]_{vuct}
=
\frac{x_{vuct} - \mu_{ct}(\bx)}{\sqrt{\sigma_{ct}^2(\bx)+\epsilon}},
\end{equation}
where
\begin{align*}
 \mu_{ct}(\bx) &= \frac{1}{HW}\sum_{vu} x_{vuct},\\
 \sigma^2_{ct}(\bx) &= \frac{1}{HW}\sum_{vu} (x_{vuct} - \mu_{ct}(\bx))^2.
\end{align*}

In practice, both batch normalization and instance normalization layers are always immediately followed by a scaling layer (Fig.~\ref{f:arch}). As discussed earlier in section~\ref{s:mdomain}, in this paper we consider either fixing the same scaling and bias parameters across domains, or make them domain specific.

\paragraph{Batch purity.} When the model is trained or tested, batches are always \emph{pure}, i.e.\ composed of data points from a single domain. This simplifies the implementation, and, most importantly, has an important effect on the BN layer. For a pure batch, BN can in fact aggressively normalize dataset-specific biases, which would not be possible for mixed batches. IN, instead, operates on an image-by-image basis, and is not affected by the choice of pure or mixed batches.

An important detail is how the BN and IN blocks are used in testing, after the network has been learned. Upon ``deploying'' the architecture for testing, the BN layers are usually \emph{removed} by fixing means and variances to fixed averages accumulated over several training batches~\cite{ioffe15batch}. Unless this is done, BN cannot be evaluated on individual images at test time; furthermore, removing BN usually slightly improves the test performance and is also slightly faster.

Dropping BN requires some care in our architecture due to the difference between pure batches from different domains $\mathcal{D}_d$. In the experiments, we test computing domain-specific means and variances $(\bm\mu_d,\bm\sigma_d^2)$, selected by a muxer from collections $(U,\Sigma)$ (Fig.~\ref{f:arch}), or share a single set of means and variances $(\bm\mu,\bm\sigma^2)$ between all domains. We also consider an alternative setting, BN+, in which BN is applied at training and test times unchanged. The disadvantage of BN+ is that it can only operate on pure batches and not single images; the advantage is that moments are estimated on-the-fly for each test batch instead of being pre-computed.

Note that the IN layer is similar to the BN+ layer in that it estimates means and variances on the fly, both at training and testing time, and applies unchanged in both phases.

\subsection{Training regime}\label{s:regime}

As noted above, training always consider pure batches. In more detail, all models are learned by means of SGD, alternating batches from each of the domain $\mathcal{D}_i$, in a round-robin fashion. This automatically balances the datasets when these have different sizes, as learning visits an equal number of training samples for each domain, regardless of the different training set sizes. This corresponds to weighing the domain-specific loss functions equally.

This design also has some practical advantages. In our implementation, different domains are assigned to different GPUs. In this case, each GPU computes the model parameter gradients with respect to a pure batch extracted from a particular dataset. Gradients are then accumulated before the descent step.

Finally, note that architectures may only partially share features, up to some depth. Obviously, domain-specific parameters are updated only from the pure batches corresponding to that domain.

\section{Experiments}\label{s:experiments}

\begin{table*}
\centering
\scalebox{0.9}{
	\begin{tabular}{l*{10}{c}}
    \toprule
  Dataset & AwA & Caltech & CIFAR10 & Daimler & GTSR & MNIST & Omniglot & Plankton & Sketches & SVHN\\
  \midrule
  \# classes & 50 & 257 & 10 & 2 & 43 & 10 & 1623 & 121 & 250 & 10 \\
  \# images  & 30k & 31k & 60k & 49k & 52k & 70k & 32k & 30k & 20k & 99k \\ 
    content & animal & object & object & pedestrian & traffic sign & digit & character & plankton & sketch & digit \\
    \bottomrule
  \end{tabular}
}
  \caption{Statistics of various datasets.}
 \label{tab:smalldatasets}
\end{table*}

\begin{figure*}[t!]
\centering
\scalebox{0.9}{
	\begin{tabular}{*{10}{C{0.077\textwidth}}}
	\includegraphics[height=0.1\textwidth,width=0.1\textwidth]{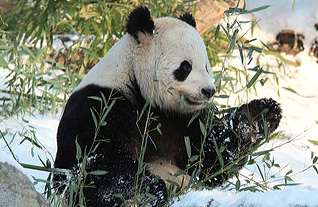}&
	\includegraphics[height=0.1\textwidth,width=0.1\textwidth]{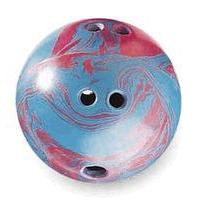}&
	\includegraphics[height=0.1\textwidth,width=0.1\textwidth]{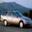}&
	\includegraphics[height=0.1\textwidth,width=0.1\textwidth]{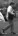}&
	\includegraphics[height=0.1\textwidth,width=0.1\textwidth]{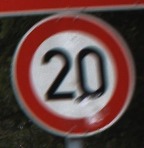}&
	\includegraphics[height=0.1\textwidth,width=0.1\textwidth]{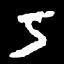}&
	\includegraphics[height=0.1\textwidth,width=0.1\textwidth]{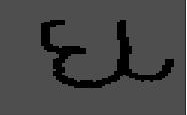}&
	\includegraphics[height=0.1\textwidth,width=0.1\textwidth]{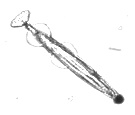}&
	\includegraphics[height=0.1\textwidth,width=0.1\textwidth]{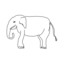}&
	\includegraphics[height=0.1\textwidth,width=0.1\textwidth]{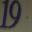}\\
	\addlinespace[-2.6ex]
	\includegraphics[height=0.1\textwidth,width=0.1\textwidth]{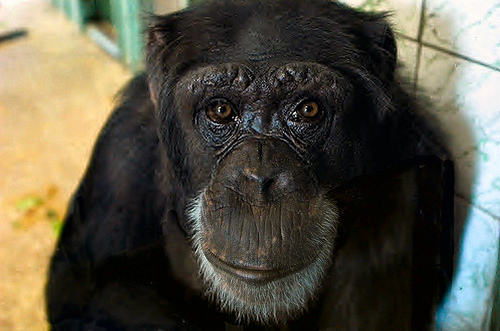}&
	\includegraphics[height=0.1\textwidth,width=0.1\textwidth]{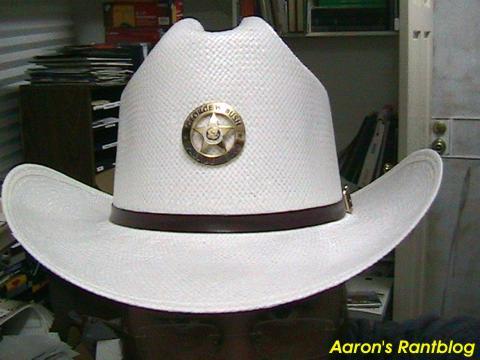}&
	\includegraphics[height=0.1\textwidth,width=0.1\textwidth]{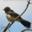}&
	\includegraphics[height=0.1\textwidth,width=0.1\textwidth]{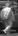}&
	\includegraphics[height=0.1\textwidth,width=0.1\textwidth]{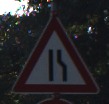}&
	\includegraphics[height=0.1\textwidth,width=0.1\textwidth]{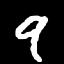}&
	\includegraphics[height=0.1\textwidth,width=0.1\textwidth]{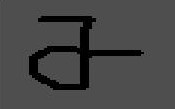}&
	\includegraphics[height=0.1\textwidth,width=0.1\textwidth]{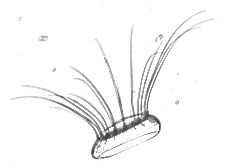}&
	\includegraphics[height=0.1\textwidth,width=0.1\textwidth]{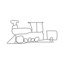}&
	\includegraphics[height=0.1\textwidth,width=0.1\textwidth]{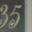}\\

	\small AwA & 	\small Caltech & 	\small CIFAR10 & 	\small Daimler & 	\small GTSR & 	\small MNIST & 	\small Omniglot & 	\small Plankton & 	\small Sketches & 	\small SVHN\\
  \end{tabular}
}
  \caption{Example images from various datasets. }
 \label{tab:samples}
\end{figure*}

\begin{table}[t]
	\centering
	\begin{tabular}{lccc}
		\toprule
		& CIFAR10 & MNIST & SVHN\\
		\midrule
		No sharing & 9.4 & 0.34 & 3.7\\
		Deep sharing & 10.2 & 0.37 & 3.7\\
		Full sharing & 10.2 & 0.38 & 3.7\\
		\bottomrule
	\end{tabular}
	\caption{Top-1 error rate (\%) for three datasets. The top row is for individually trained networks per dataset. Deep sharing corresponds to sharing all the convolutional but the last classifier layer. Full sharing corresponds to sharing all parameters including \emph{final classifier} parameters with domain-specific scale and bias parameters. Note that all three datasets have ten classes and this allows us to share classifier parameters.}
	\label{tab:shareclassifier}
\end{table}

Experiments focus on image classification problems in two scenarios. In the first one, different architectures and learning strategies are evaluated on a portfolio of 10 very diverse image classification datasets (section~\ref{s:small}), from planktons to street numbers. For computational reasons, these experiments consider relatively small $64\times 64$ pixels images and tens of thousands training images per domain. In the second scenario, we test similar ideas on larger datasets, including ImageNet, but in a less extensive manner due to the computational cost (section~\ref{s:large-data}).

\subsection{Small datasets}\label{s:small}

\paragraph{Data.} We choose 10 image classification tasks from very diverse domains including objects, hand-written digit and characters, pedestrians, sketches, traffic signs, planktons and house numbers. The dataset statistics are summarized in~Table~\ref{tab:smalldatasets} and a few example images are given in~Table~\ref{tab:samples}.

In more detail, \textbf{Animals with Attributes (AwA)}~\cite{lampert2009learning} contains 30475 images of 50 animal species. While the dataset is introduced for zero-shot learning, it provides class labels for each image. \textbf{Caltech-256}~\cite{griffin2007caltech} is a standard object classification benchmark that consists of 256 object categories and an additional background class. \textbf{CIFAR10}~\cite{krizhevsky2009learning} consists of 60000 $32\times32$ colour object classes in 10 classes. \textbf{Daimler Mono Pedestrian Classification Benchmark}~\cite{munder2006experimental} contains a collection of pedestrian and non-pedestrian images. Pedestrians are cropped and resized to $18\times36$ pixels. \textbf{The German Traffic Sign Recognition (GTSR) Benchmark}~\cite{Stallkamp2012} contains cropped images of 43 traffic signs. Sizes of the traffic signs vary between $15\times15$ and $222\times193$ pixels. \textbf{MNIST}~\cite{lecun1998mnist} contains 70000 handwritten digits which are centred in $28\times28$ images. \textbf{Omniglot}~\cite{lake2015human} consists of 1623 different handwritten characters from 50 different alphabets. The dataset is originally designed for one shot learning. Instead we include all the character categories in train and test time. \textbf{Plankton imagery data}~\cite{planktonv1} is a classification benchmark that contains 30336 images of various organisms ranging from the smallest single-celled protists to copepods, larval fish, and larger jellies. \textbf{Human Sketch dataset}~\cite{eitz2012hdhso} contains 20000 human sketches of every day objects such as ``book'', ``car'', ``house'', ``sun''. \textbf{The Street View House Numbers (SVHN)}~\cite{netzer2011reading} is a real-world digit recognition dataset with around 70,000 images which are centred around a single character and resized into $32\times32$ pixels.

As the majority of datasets differ in terms of image resolutions and characteristics, images are resized to $64\times64$ pixels, greyscale ones are converted into RGB by setting the three channels to the same value. Though it would be possible to maintain the images in the original resolution, using a single scale simplifies the network design. Each dataset is also whitened, by subtracting its mean and dividing it by its standard deviation per channel. For the datasets that do not have a fixed train and test splits, we use $80\%$ to $20\%$ ratio for train and test data respectively.

\paragraph{Architectures.} We choose to use the state-of-the-art Residual Networks \cite{he2016identity} due to their remarkable capacity and performance. More specifically, for this experiment we select the ResNet-38 model. This network has a stack of 4 residual units with $3\times3$ convolutions for each feature map size ($\{64,32,16,8\}$) and with number of filters $\{16, 32, 128, 256\}$ respectively. The network ends with a global average pooling layer and a fully connected layer followed by softmax for classification. As the majority of the datasets have a different number of classes, we use a dataset-specific fully connected layer in our experiments unless otherwise stated.

As explained in section~\ref{s:regime}, datasets are balanced by sampling batches from different ones in a round-robin fashion during training. We follow the same data augmentation strategy in~\cite{he2016identity}, the $64\times64$ size whitened image is padded with 8 pixels on all sides and a $64\times64$ patch randomly sampled from the padded image or its horizontal flip. Note that as MNIST, Omniglot and SVHN contain digits and characters, we do not augment flipped images from these datasets. The networks are trained using stochastic gradient descent with momentum. The learning range is set to 0.1 and gradually reduced to 0.0001 after a short warm-up training with a learning rate of 0.01 as in~\cite{he2016identity}. The weight decay and momentum are set to 0.9 and 0.0001 respectively. In test time, we use only the central crop of images and report percentages of top-1 error rate.

Next, we experiment with sharing features up to different depths in the architectures. To this end, the ResNet-38 model $\phi_d = \phi_d' \circ \phi_0$ is modified to have a branch $\phi'_d$ for each task $\mathcal{D}$, stemming from a common trunk $\phi_0$.

\paragraph{Baseline: no sharing.} As a baseline, a different ResNet-38 model (i.e.\ $\phi'_d = \phi_d$) is trained from scratch for each dataset until convergence (25k iterations are sufficient) using a batch size of 128 images and BN. Although our focus is not obtaining state-of-the-art results but demonstrating the effectiveness of sharing a representation among different domains, the chosen CNN provides a good speed and performance trade-off and achieves comparable results to state-of-the-art methods (see Table~\ref{tab:multismall}). The much deeper ResNet-1001~\cite{he2016identity} obtains $4.62\%$ error rate in CIFAR-10 (compared to our $9.4\%$), DropConnect~\cite{wan2013regularization} with a heavier multi-column network obtains $0.21\%$ in MNIST (compared to our $0.3\%$) and Lee \etal~\cite{lee2016generalizing} report $1.69\%$ (compared to our $3.7\%$) in SVHN by using more sophisticated pooling mechanisms. While the network yields relatively low error rates in the majority of the datasets, the absolute performance is less good in AwA and Caltech256. However, this is inline with the results reported in the literature, where good  performance on such datasets was shown to require pre-training on a very large dataset such as ImageNet~\cite{Donahue2013,zeiler14visualizing}. In short, this validates our ResNet-38 baseine as a good and representative architecture.

\begin{table*}
\centering
\scalebox{0.85}{
  \begin{tabular}{l*{10}{r}|r}

    \toprule
      Sharing & AwA & Caltech & CIFAR10 & Daimler & GTSR & MNIST & Omniglot & Plankton & Sketches & SVHN & mean\\
  	
    \midrule
    no sharing              & 77.9 & 85.1 &  9.4 & 10.4 &  4.2 &  0.34 &  13  & 25.7 &  31.1 &  3.7 & 26.1\\ 
    \midrule
    deep & 73.7 & 82.6 & 11.7 &  5.2 &  \bf{3.5} &  0.38 & 13.1 & \bf{24.9} & 30.8 & 4.1 & 25.0\\  	
    \midrule
    partial (block 1-3)  & 76.6 & 84.0 & 12.4 & 7.0  & 4.0  & \bf{0.29}  &  13.8 & 26.2 & 33.1 & 4.4  & 26.2\\
    partial (block 2-4)  & 76.6 & 84.0 &  9.9 & 7.7  & 3.3  & 0.49  &  12.1 & 25.3 & 30.7 & \bf{3.5}  & 25.3\\
    \midrule
    deep ($\times2$ params) & 74.0 & 81.7 &  9.1 &  4.2 &  4.1 &  0.43 & 12.3 & 25.8 & 29.1 &  \bf{3.5} & 24.4\\
    deep ($\times4$ params) & \bf{73.1} & \bf{81.5} & \bf{7.2}  & \bf{4.1}  & \bf{3.5}  & 0.35  & \bf{11.8} & 25   & \bf{27.6} & \bf{3.5}  & \bf{23.8}\\
    \bottomrule
  \end{tabular}
	}
  \caption{Top-1 error rate(\%) for various tasks. The table shows the results in case of no feature sharing between different domains (first row), deep feature sharing of all convolutional weights (deep), partial sharing for selected convolutional weights in block 1-3 and block 2-4 and deep sharing with more convolutional filters ($\times2$ and $4\times$ number of filters). }
 \label{tab:multismall}

\end{table*}

\paragraph{Full sharing.} Next, we consider the opposite of no sharing and share \emph{all} the parameters of the network (i.e.\ $\phi_d = \phi_0$). A common belief is that only the relatively shallow layers of a CNN are shareable, whereas deeper ones are more domain-specific~\cite{lenc15understanding}. Full sharing challenges this notion.

In this experiment, ResNet-38 is configured with BN with domain-specific scaling parameters $(\bm s_d, \bm b_d)$ and moments $(\bm \mu_d,\bm\sigma_d^2)$.  A single CNN is trained on three domains, CIFAR, MNIST, and SVHN, because such domains happens to contain exactly 10 classes each. Although CIFAR objects and MNIST/SVHN digits have nothing in common, we randomly pair digits with objects. This allows to share  all filter parameters, including the final classification layer, realising full sharing. 

As shown in~Table~\ref{tab:shareclassifier}, evaluated on the different datasets, the performance of this network is nearly the same as learning three independent models. This surprising result means that the model has sufficient capacity to learn classifiers that respond strongly either to a digit in MNIST or SVHN, or to an object in CIFAR, essentially learning an \emph{or} operator. The question then is whether combining more problems together can eventually exceed the capacity of the network.

\paragraph{Deep sharing.} Next, we experiment with sharing all layers except the last one, which  performs classification. In this case, therefore, $\phi'_d$ is a single convolutional layer and $\phi_0$ contains the rest of the network, including all but the last fully connected layer. This setup is similar to full sharing, but allows to combine classification tasks even when these do not have an equal number of classes.

The results in~Table~\ref{tab:multismall} and~Table~\ref{tab:shareclassifier} show that the shared CNN performs \emph{better} than training domain-specific networks. Remarkably, this is true for all the tasks, and reduces the average error rate by $1\%$. Remarkably, this improvement is obtained while reducing the overall number of parameters by a factor of 10.

\paragraph{Partial sharing.} Here, we investigate whether there can be a benefit in specializing at least part of the shared model for individual tasks. We test two settings. In the first setting, the network has dataset-specific parameters in the shallower block (i.e.\ the first stack of 4 residual units) --- this should be beneficial to compensate for different low-level statistics in the domains.  In the second setting, instead, the network specializes the last block--- this should be beneficial in order to capture different higher-level concepts in for different tasks. Interestingly, the results in~Table~\ref{tab:multismall} show that deep sharing is, for this choice of datasets and model, the best configuration. Specializing the last block is only marginally worse ($-0.3\%$) and better ($+0.9\%$) than specializing the first block. This may indicate that high-level specialization is preferable.

\paragraph{Network capacity.} Experiments so far suggested that the model has sufficient capacity to accommodate all the tasks, despite their significant diversity. In fact, ten individual networks perform worse than a single, shared one. Next, we increase the capacity of the model, but we keep sharing all such parameters between tasks. In order to do so, we increase the number of convolutional filters twice ($\{64, 128, 256, 512\}$) and four times ($\{128, 256, 512, 1024\}$), which increases the number of parameters 4 and 16 times. Differently from learning 10 independent networks, this setup allows the model to better use the added capacity to accommodate the different tasks, reducing the mean error rate by $0.6\%$ and $1.2\%$ points, respectively. The fact that joint training can exploit the added capacity better suggests that the different domains overlap synergistically, despite their apparent differences.

\paragraph{Normalization strategies.} So far, we have shown that learning a single CNN for the 10 domains is not only possible, but in fact preferable to learining individual models. However, this CNN used a specific normalization strategy, BN, as well as domain-specific scaling parameters $(\bm s_d, \bm b_d)$ and moments $(\bm\mu_d,\bm\sigma_d^2)$.

In Table~\ref{tab:norm} we examine the importance of these design decisions. First, we note that BN with domain-agnostic scaling $(\bm s, \bm b)$ and moments $(\bm\mu_d,\bm\sigma^2)$ performs very poorly on the test set, comparable to random chance, clearly due to its inability to compensate for the large variance among the different domains. If BN is applied at test time (BN+), such that moments are computed on the fly but domain-agnostic scaling is still used, results are better but still poor ($46.3\%$). Domain-agnostic scaling works well as long as at least the moments are domain-specific ($27.3\%$). However, the best combination is to use domain-specific moments and scalings ($25\%$).

In contrast to BN, IN, which normalizes images individually, works just as well with domain-specific and domain-agnostic scaling. The price to pay is a $5\%$ drop in performance compared to BN with domain-specific parameters. However, this strategy has a significant practical advantage: IN with domain-agnostic scaling effectively uses only a single set of parameters, including all filter weights and scaling factors, for all domains. This suggest that such a representation may be applicable to novel domain without requiring any domain-specific tuning of the normalization parameters at all.

\begin{table}
	\centering
	\begin{tabular}{lllr}
		\toprule
		normalization & $(\bm s,\bm b)$ & $(\bm\mu,\bm\sigma)$ & mean error\\
		\midrule
		BN         & universal & universal & --   \\
		BN+        & universal & --        & 46.3 \\
	    BN         & universal & domain    & 27.3 \\ 
		BN         & domain    & domain    & \bf{25} \\
		IN         & universal & --        & 30.2 \\
	    IN         & domain    & --        & 30.4 \\
	    \bottomrule
	\end{tabular}
	\caption{Mean top-1 error over the 10 datasets for different normalization strategies and domain-specific or domain-agnostic (universal) choice of the scaling factors $(\bm s,\bm b)$ and BN moments $(\bm\mu,\bm\sigma)$. BN+ corresponds to applying BN at test time as well, which does not use pre-computed moments.}
	\label{tab:norm}
\end{table}

\subsection{Large datasets}\label{s:large-data}

\begin{figure}
\centering
\scalebox{0.9}{
	\begin{tabular}{*{3}{c}}
	\includegraphics[width=0.2\textwidth]{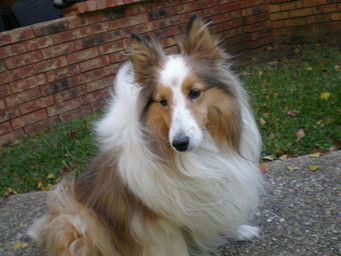}&
	\includegraphics[height=0.1\textwidth]{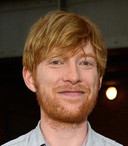}&
	\includegraphics[width=0.1\textwidth]{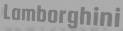}\\
	\addlinespace[-2.6ex]
	\includegraphics[width=0.2\textwidth]{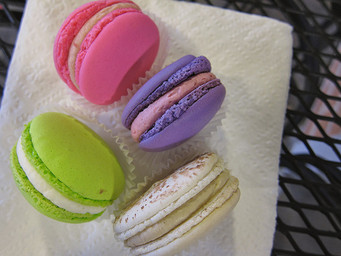}&
	\includegraphics[height=0.1\textwidth]{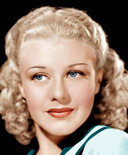}&
	\includegraphics[width=0.1\textwidth]{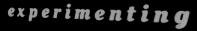}\\
	\small ImageNet & 	\small VGG-Face & \small Synth90k\\
  \end{tabular}
}
  \caption{Example images from the large-scale datasets are shown in their relative sizes.}
 \label{tab:samples-large}
\end{figure}

\paragraph{Data.} In this part, we consider three large scale computer vision tasks: object classification in ImageNet~\cite{russakovsky14imagenet}, face identification in VGG-Face~\cite{parkhi15deep}, and word classification in Synth90k~\cite{jaderberg14synthetic} dataset (Table~\ref{tab:largedatasets}). ImageNet contains 1000 object categories and 1.2 million images. VGG-Face dataset consists of 2.6 million face images of 2622 different people which are centered and resized into a fixed height of 128 pixels and a variable width. The Synth90k dataset contains approximately 9 million synthetic images for a 90k word lexicon which are generated with a fixed height of 32 pixels and variable width. We show example images from these datasets in Table~\ref{tab:samples-large}.

\paragraph{Implementation details.} ImageNet images are resized to $256$ pixels on their shortest side and maintaining the original aspect ratio. During training, random $224\times224$, $112\times56$ and $32\times128$ patches are cropped from ImageNet, VGG-Face and Synth90k  respectively. As different input sizes lead to different feature map sizes at the last convolutional layer (a $6\times 6$ map for ImageNet down to a tiny $1 \times 6$ map for the smallest Synth90k images -- see below), we share the convolutional feature maps among the three tasks but use domain-specific fully connected layers. In training, we augment the data by randomly cropping, flipping and varying aspect ratio with the exception that we do not flip the images from Synth90k as they contain words. At test time, we only use a single center crop and report top1-error rates. The best normalization strategy (BN with domain specific scaling) identified in~\ref{s:small} is used.

\paragraph{Results.} We conduct two experiments. In the first one, a network is trained simultaneously on the ImageNet and VGG-Face datasets, with a significant difference in content as well as resolution between domains. We use an AlexNet model~\cite{krizhevsky12imagenet}, adapt the dimensionality of the first fully connected layer (\texttt{fc6}) for the VGG-Face dataset, and train the networks from scratch. For this setting, the baseline is obtained by training two individual networks without any parameter sharing, obtaining $40.5\%$ and $25.7\%$ top-1 error rates on the ImageNet and VGG-Face respectively (see~Table~\ref{tab:largedatasets} --- this is the same as published results). Sharing the convolutional weights between these tasks achieve comparable performance (there is a marginal drop of $~1\%$ accuracy), illustrating once more the high degree of shareability of such representations.

In the second setting, we push the envelope by adding the Synth90k dataset which contains synthetically generated words for 90k different word classes. For this experiment, we use the higher-capacity model VGG-M-128 from~\cite{chatfield14return}. This model has only 128 filters in the second to last fully connected layer (\texttt{fc7}), instead of 4096. As the Synth90k dataset contains 90k classes, having a small 128-dimensional bottleneck is necessary in order to maintain the size  of the 90k classes classifier matrix (which is $128 \times 90k$) reasonable. Since Synth90k images are much smaller than the other two datasets, the last downsampling layer ({\tt pool5}) is not used for this domain.

Without parameter sharing, this network performs similarly to AlexNet (due to the bottleneck which partially offsets the higher capacity). As before, the convolutional layer parameters are shared, and the fully-connected layers parameters are not. The results (see the bottom row in~Table~\ref{tab:largedatasets}) show that the capacity of the model is pushed to its limit. Performance on ImageNet and VGG-Face is still very good, with a minor hit of 2-3\%, but there is a larger drop for Synth90k ($26.9\%$ error). Note that the total number of parameters in the joint network is a third of the sum of the individual network parameters. In order to have a fair comparison, we evaluate the performance of three independent models with a third of the parameters each (see the $4^\text{th}$ row in~Table~\ref{tab:largedatasets}). We show that the jointly trained model performs dramatically better than the individual models, despite the fact that the total number of parameters is the same.

\begin{table}
  \centering
  \begin{tabular}{lrrr}
	  \toprule
    & ImageNet & VGG-Face & Synth90k\\
    \midrule
  No sharing & \bf{40.5} & \bf{25.7} & -\\
  Deep (conv1-5) & 41.8 & 26.3 & -\\
  \midrule
  No sharing & \bf{40.8} & \bf{25.7} & \bf{12.1}\\
  No sharing ($^1/_3$) & 49.2 & 57.1 & 39.0\\
  Deep (conv1-5) & 43.0 & 27.8 & 26.8\\
  \bottomrule
  \end{tabular}
  \caption{Top-1 error rate (\%). The first and second rows show the results for the AlexNet without and with sharing parameters on ImageNet and VGG-Face datasets respectively. The third and fourth depict the results for a VGG-M-128 without parameter sharing in different capacities. $^1/_3$ indicates the number of parameters reduced to a thir in each individual network. The last one show the results for the same network with sharing parameters on ImageNet, VGG-Face and MJSynth datasets respectively. }
  \label{tab:largedatasets}
\end{table}

\section{Conclusions}\label{s:conclusions}

As machine vision consolidates, the challenge of developing universal models that, similarly to human vision, can be trained once and solve a large variety of problems, will come into focus. A component of such systems will be \emph{universal representations}, i.e.\ feature extractors that work well for all visual domains, despite the significant diversity of the latter.

In this paper, we have shown that standard deep neural networks are already capable of learning very different visual domains together, with a high degree of information sharing. However, we have also shown that successful sharing requires tuning certain normalization parameters in the networks, preferably by using domain-specific scaling factors, in order to compensate for inter-domain statistical shifts. Alternatively, techniques such as instance normalization can compensate for such difference on the fly, in a domain-agnostic manner.

Overall, our findings are very encouraging. Universal representations seem to be within the grasp of current technology, at least for a wide array of real-world problems. In fact, while our most convincing results have been obtained for smaller datasets, we believe that larger problems can be addressed just as successfully by a moderate increase of model capacity and other refinements to the technology.

\small
\bibliographystyle{ieee}
\bibliography{refs,bibliography/bibliography}
\end{document}